\documentclass[10pt,conference]{IEEEtran}
\usepackage{cite}
\usepackage{amsmath,amssymb,amsfonts}
\usepackage{algorithm}
\usepackage{algorithmic}
\usepackage{pgfplots}
\pgfplotsset{compat=1.18}
\usepackage{graphicx}
\usepackage{textcomp}
\usepackage{xcolor}
\usepackage{booktabs}
\usepackage{multirow}
\usepackage{hyperref}
\usepackage{cleveref}
\usepackage{subcaption}

\def\BibTeX{{\rm B\kern-.05em{\sc i\kern-.025em b}\kern-.08em
    T\kern-.1667em\lower.7ex\hbox{E}\kern-.125emX}}

\title{PACE: Adaptive Budget Allocation for Time-Efficient Embodied Planning}

\author{
    \IEEEauthorblockN{Yuchen Huang, Xijiang Ying, Zhenhua Ma,}
    \IEEEauthorblockN{Xiaxiang Yuan, Zhijie Gao, Jiayi Huang,}
    \IEEEauthorblockN{Ruichi Mao, Jiazheng Zhang, Hongsheng Ti,}
    \IEEEauthorblockN{Maotao Tian, Rong Shi, Lu Zhao, Shizhuang Zhang,}
    \IEEEauthorblockN{Zhuo Cui, He Wang, Ling Liu, and Wei Zhang$^{*}$}
    \IEEEauthorblockA{\textit{ZTE Corporation}\\
    Shanghai, China\\
    $^{*}$Corresponding author: zhang.wei14@zte.com.cn}
}

\begin{document}

\maketitle

\begin{abstract}
Reasoning-enhanced large language models have achieved remarkable improvements in planning tasks, yet their deployment in embodied systems remains impractical due to prohibitive inference delays—often exceeding minutes per planning instance. The fundamental bottleneck stems from the serial nature of existing paradigms: models must complete all reasoning before any action execution, leaving execution time windows entirely unexploited. We introduce PACE (Planning with Adaptive Cognitive Effort), a framework that enables interleaved reasoning and execution through two key innovations: an Interleaved Think-Act architecture that pipelines cognitive processing with action execution, and a Dynamic Budget Allocator that adapts reasoning token budgets to available execution time windows. On the Robotouille benchmark \cite{gonzalez-pumariega2025robotouille} using Qwen3-8B-AWQ, PACE achieves a 10\% success rate—representing a 67\% improvement over the ReAct+Think baseline—while delivering 6.9$\times$ acceleration in thinking time compared to unconstrained reasoning. The framework hides 66.8\% of thinking time within execution windows, demonstrating that strategic cognitive effort allocation can simultaneously improve both planning quality and time efficiency. These results provide evidence that time-aware architectural innovations enable reasoning models to operate in latency-sensitive embodied domains where they were previously impractical.
\end{abstract}


\section{Introduction}
\label{sec:introduction}

The emergence of reasoning-enhanced large language models has fundamentally transformed the landscape of AI planning capabilities. Models such as OpenAI's o1 and DeepSeek-R1 leverage extended chain-of-thought processing to achieve substantial improvements on benchmarks like PlanBench, with reported gains exceeding 60\% in success rates compared to their non-reasoning counterparts \cite{openai2023technical,bubeck2023sparks}. These advances stem from the models' ability to engage in deliberate, multi-step reasoning before committing to actions—a paradigm that mirrors human cognitive processes of careful deliberation before execution. However, this breakthrough comes at a severe cost: reasoning models generate hundreds or thousands of thinking tokens before producing actionable outputs, resulting in inference latencies that render them impractical for real-time robotic systems \cite{kasneci2023chatgpt}.

The deployment of reasoning models in embodied planning contexts exposes three critical bottlenecks that existing approaches fail to address. First, inference delays have become unacceptable for interactive systems—planning a single task instance can consume minutes of computation time, during which a robot remains idle and unresponsive. Second, the prevailing paradigm enforces strict seriality between thinking and acting \cite{guo2023plan,wang2023language}: models must complete their entire reasoning process before any physical action begins. This stands in stark contrast to human cognition, where chefs chop vegetables while steaks sear and mechanics retrieve tools while diagnostics run. Third, reasoning budgets remain static and globally configured, unable to adapt to the varying cognitive demands of different planning steps or the temporal constraints imposed by execution windows \cite{chen2023milestones,gangi2023deep}. A simple pick-and-place operation receives the same reasoning allocation as a complex multi-step assembly sequence, leading to both computational waste and critical reasoning gaps.

Prior work on LLM-based planning has made substantial progress in improving success rates through iterative reasoning-acting loops, hierarchical task decomposition \cite{ajay2023compositional}, and structured prompting strategies. The ReAct paradigm demonstrated that interleaving reasoning traces with action execution enables more robust planning through environmental feedback. However, these approaches treat time as a secondary concern—if considered at all—focusing primarily on solution quality while ignoring the temporal constraints that govern real-world deployment \cite{garrett2020pddlstream,huang2022inner}. Recent investigations into budget-constrained reasoning have explored token limitation strategies \cite{chen2023milestones,gangi2023deep}, yet these methods apply uniform constraints across all planning steps without accounting for the natural temporal structure of embodied tasks, where action execution creates predictable windows for concurrent cognitive processing.

We present PACE (Planning with Adaptive Cognitive Effort), a framework that bridges the gap between reasoning model capabilities and real-time embodied planning requirements. PACE introduces two core innovations that work synergistically to optimize the success-time trade-off. The Interleaved Think-Act (ITA) architecture enables reasoning processes to execute concurrently with action execution, exploiting the temporal structure of embodied tasks where physical operations create natural windows for cognitive processing. The Dynamic Budget Allocator (DBA) adapts thinking token budgets to the available execution time of each action, scaling cognitive effort to match temporal constraints while prioritizing budget allocation for critical decision points. Together, these components enable reasoning models to achieve higher success rates while dramatically reducing wall-clock planning time.

The contributions of this work are fourfold:
\begin{enumerate}
    \item We formalize the Time-Delay-Aware Embodied Planning (TDAEP) problem, introducing a pipeline time model that captures the opportunity for concurrent reasoning during action execution.
    \item We propose the PACE framework comprising the ITA architecture for concurrent think-act processing and the DBA mechanism for execution-aware budget allocation.
    \item We develop a hybrid budget control mechanism combining prompt-based guidance with token truncation, enabling precise reasoning control without model retraining.
    \item We establish a Pareto efficiency evaluation framework for embodied planning that jointly optimizes success rate and total completion time, demonstrating PACE's dominance over existing paradigms.
\end{enumerate}

\section{Related Work}
\label{sec:related}

\subsection{LLM Planning Paradigms}

Large language models have emerged as powerful planners across diverse domains, with several paradigms developed to leverage their reasoning capabilities \cite{huang2022inner,valmeekam2022large}. The Input-Output (IO) paradigm represents the simplest approach, where models generate complete plans in a single inference step without intermediate feedback. While computationally efficient, IO planning lacks the ability to correct errors or adapt to unexpected environmental states, limiting its effectiveness in complex embodied scenarios \cite{singh2023progprompt}. The ReAct paradigm addresses this limitation by interleaving reasoning traces with action execution, enabling models to incorporate environmental feedback into subsequent planning steps. This iterative approach has demonstrated substantial improvements in planning accuracy, particularly for tasks requiring multi-step reasoning with environmental interaction \cite{guo2023plan}.

Recent advances in reasoning models have introduced dedicated thinking phases that precede action generation. Models like GPT-4 with reasoning mode and DeepSeek-R1 allocate extended computation to deliberate thinking before committing to actions, achieving significant improvements on planning benchmarks \cite{openai2023technical}. However, these reasoning-enhanced approaches exacerbate the temporal bottleneck: the extended thinking phases can consume orders of magnitude more inference time than standard approaches, making them impractical for time-critical embodied applications \cite{kasneci2023chatgpt}. Our work addresses this fundamental tension by enabling reasoning processes to overlap with action execution, preserving the benefits of deliberate thinking while mitigating temporal costs.

\subsection{Budget-Constrained Reasoning}

The computational cost of reasoning models has motivated research into budget allocation and constraint mechanisms \cite{chen2023milestones,gangi2023deep}. Prior work has explored static budget constraints that limit total thinking tokens across all planning steps, demonstrating that moderate constraints can paradoxically improve performance by preventing ``overthinking''—the generation of verbose but unhelpful reasoning chains. Dynamic budget approaches have been proposed for specific contexts, adjusting reasoning depth based on task complexity or uncertainty estimates \cite{gangi2023deep}. However, existing methods fail to consider the temporal structure of embodied tasks, where action execution creates predictable windows for cognitive processing.

Cognitive science research on effort allocation provides theoretical grounding for adaptive budget strategies. Studies have shown that humans dynamically adjust cognitive effort based on expected value, task difficulty, and available time \cite{westbrook2020dopamine,clay2022rewarding}. The prefrontal cortex plays a critical role in this adaptive control, modulating cognitive resources based on task demands \cite{friedman2021role,bogdanov2021cognitive}. Our DBA mechanism draws inspiration from these findings, implementing a difficulty-aware coefficient that scales reasoning budgets to match both temporal constraints and cognitive demands. Unlike prior approaches that apply uniform budgets or simple heuristics, PACE integrates execution time awareness into the budget allocation process, enabling principled reasoning control that respects real-time constraints \cite{parr2023cognitive,jiang2024adaptive}.

\subsection{Embodied Planning with LLMs}

The integration of LLMs into robotic and embodied systems has attracted substantial research attention. Prior work has demonstrated LLM-based task planning for household robots \cite{singh2023progprompt}, manufacturing systems \cite{bolu2021adaptive}, and autonomous navigation \cite{zhong2020hybrid}. These systems typically employ hierarchical planning architectures where LLMs handle high-level task decomposition while specialized controllers manage low-level execution \cite{ajay2023compositional}. However, the temporal dimension of planning remains underexplored in this literature—most evaluations focus on success rates and plan quality while treating planning time as a secondary concern \cite{valmeekam2022large}.

The Robotouille benchmark \cite{gonzalez-pumariega2025robotouille} represents a significant step toward standardized evaluation of LLM planning in embodied contexts, providing a kitchen-based task environment with both synchronous and asynchronous execution modes. Our experiments leverage this benchmark to evaluate PACE against established baselines, demonstrating that time-aware planning can achieve superior success rates while dramatically reducing total completion time. The benchmark's action time model—where different operations have distinct execution durations—provides the temporal structure that PACE exploits for concurrent reasoning.

\subsection{Parallel and Concurrent Processing in Planning}

Parallel processing has a rich history in classical AI planning, with algorithms designed to exploit problem structure for concurrent computation \cite{strub2020adaptively,gochev2021path}. Prior work has explored parallel state-space search, distributed heuristic evaluation, and concurrent plan refinement. However, these approaches focus on computational parallelism within the planning algorithm itself, rather than the temporal relationship between planning and execution. In embodied systems, the sequential dependency between action selection and execution has been treated as an immutable constraint—our work challenges this assumption by enabling cognitive processing during execution windows.

Recent work on adaptive planning has explored context-sensitive strategy selection \cite{choudhury2020adaptive} and real-time plan adjustment \cite{bolu2021adaptive}. These approaches adapt planning behavior to environmental conditions but do not address the fundamental seriality between thinking and acting. PACE introduces a novel form of temporal adaptation: rather than modifying planning strategies based on environmental feedback, we modify the temporal structure of the planning-execution loop itself, enabling reasoning to proceed concurrently with action execution.

\section{Problem Formulation}
\label{sec:problem}

\subsection{Traditional Serial Planning Model}

We consider embodied planning within the Markov Decision Process (MDP) framework $\mathcal{M} = \langle \mathcal{S}, \mathcal{A}, \mathcal{T}, \mathcal{R} \rangle$, where $\mathcal{S}$ denotes the state space, $\mathcal{A}$ the action space, $\mathcal{T}$ the transition function, and $\mathcal{R}$ the reward function. Traditional LLM planning paradigms operate under a serial time model where reasoning and execution proceed sequentially:

\begin{equation}
T_{\text{total}}^{\text{serial}} = \sum_{i=0}^{N} \left( t_{\text{think}}^{(i)} + t_{\text{exec}}^{(i)} \right)
\label{eq:serial}
\end{equation}

where $t_{\text{think}}^{(i)}$ represents the LLM inference time for step $i$, $t_{\text{exec}}^{(i)}$ the physical execution time, and $N$ the total number of steps. This model captures the fundamental inefficiency of existing paradigms: even when the robot executes actions lasting several seconds, the LLM remains idle, unable to begin reasoning for the next step until execution completes. Our empirical analysis reveals that reasoning time dominates total time in reasoning-enhanced models, accounting for over 90\% of the planning duration in the ReAct+Think configuration.

\subsection{Time-Delay-Aware Embodied Planning}

We introduce the Time-Delay-Aware Embodied Planning (TDAEP) problem, which formalizes the opportunity for concurrent reasoning during action execution. The key insight is that when action $a_{i-1}$ executes over duration $t_{\text{exec}}^{(i-1)}$, the LLM can simultaneously reason about action $a_i$. This pipeline structure yields a modified time model:

\begin{equation}
T_{\text{total}}^{\text{pipe}} = t_{\text{think}}^{(0)} + \sum_{i=1}^{N} \max\left(t_{\text{think}}^{(i)}, t_{\text{exec}}^{(i-1)}\right) + t_{\text{exec}}^{(N)}
\label{eq:pipeline}
\end{equation}

The pipeline model exhibits three important properties. First, the serial model upper-bounds the pipeline model: $T_{\text{total}}^{\text{pipe}} \leq T_{\text{total}}^{\text{serial}}$, with equality only when no overlap is possible. Second, when all thinking fits within execution windows ($t_{\text{think}}^{(i)} \leq t_{\text{exec}}^{(i-1)}$ for all $i \geq 1$), reasoning overhead reduces to just the initial thinking time. Third, the optimization space—the time savings achievable through pipelining—is quantified by $\sum_{i=1}^{N} \min(t_{\text{think}}^{(i)}, t_{\text{exec}}^{(i-1)})$, representing the thinking time hidden within execution windows.

\subsection{Action Execution Time Model}

Different actions have distinct execution durations based on physical requirements. We define the execution time function:

\begin{equation}
\!\!\!t_{\text{exec}}(a) \!=\! \begin{cases}
t_{\text{move}} & \texttt{Move} \\
t_{\text{manip}} & \texttt{Pick, Place, Stack, Unstack} \\
t_{\text{process}} & \texttt{Cut, Cook, Fry, Boil} \\
t_{\text{noop}} & \texttt{Do nothing}
\end{cases}
\!\!\!
\label{eq:texec}
\end{equation}

Based on realistic robotic operation times, we parameterize $t_{\text{move}} = 3.0$\,s for navigation, $t_{\text{manipulate}} = 4.0$\,s for manipulation actions, and $t_{\text{process}} = 5.0$\,s for processing operations. These values provide the temporal structure that PACE exploits for concurrent reasoning.

\subsection{Pareto Optimality in Embodied Planning}

We evaluate planning strategies along two dimensions: success rate $\text{SR}(\pi)$ and average total time $\overline{T}(\pi)$. A strategy $\pi^*$ is Pareto optimal if no other strategy achieves both higher success rate and lower time. This formulation captures the practical reality of embodied deployment: systems must balance planning quality against response time, and the optimal operating point depends on application-specific requirements. PACE aims to advance the Pareto frontier by achieving higher success rates than existing methods while simultaneously reducing total planning time.

\section{PACE Framework}
\label{sec:method}

\subsection{Overall Architecture}

PACE comprises three integrated components that work together to enable time-efficient embodied planning. The Interleaved Think-Act (ITA) loop restructures the planning-execution relationship to enable concurrent processing. The Dynamic Budget Allocator (DBA) computes appropriate thinking token budgets based on available execution time windows and task difficulty. The Hybrid Budget Control mechanism ensures reasoning adheres to allocated budgets through a combination of prompt guidance and token truncation. These components operate within a reasoning-mode LLM, such as DeepSeek-R1 or Qwen3 with thinking enabled, leveraging the model's inherent chain-of-thought capabilities while imposing temporal constraints.

\begin{figure*}[!t]
\centering
\includegraphics[width=0.98\textwidth, height=0.45\textheight, keepaspectratio]{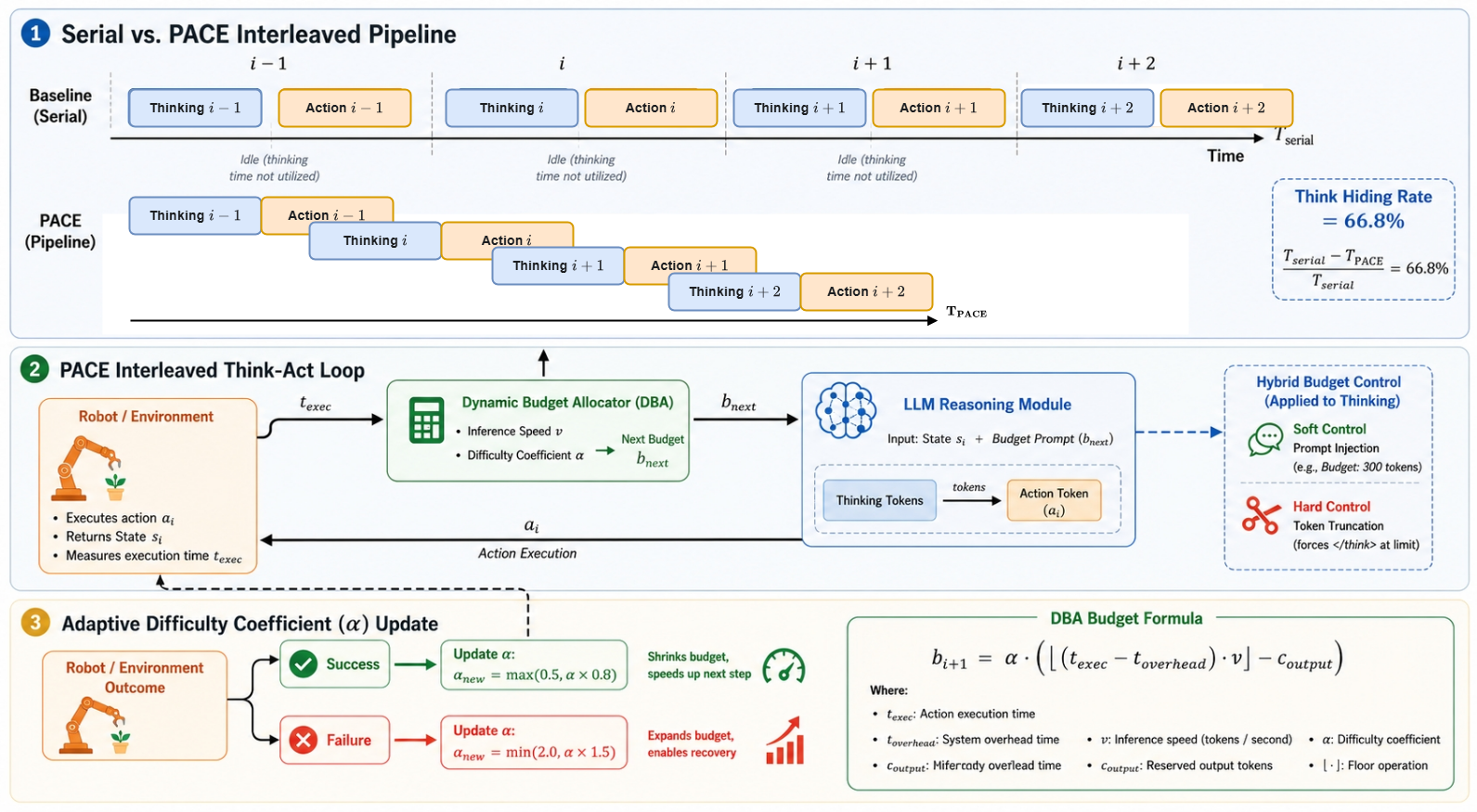}
\caption{Overview of the PACE framework. The system interleaves thinking and acting through the ITA loop (top), adapts cognitive budgets via the DBA based on execution time and difficulty (left), and ensures temporal constraints through hybrid budget control (right). The difficulty coefficient $\alpha$ dynamically scales thinking tokens after each action based on success/failure feedback.}
\label{fig:overview}
\end{figure*}

The framework maintains a planning loop where each iteration consists of three phases. In the reasoning phase, the LLM generates thinking tokens up to the allocated budget while the previous action executes concurrently. In the action phase, the selected action begins execution, and the DBA computes the budget for the next reasoning phase based on the expected execution duration. In the update phase, environmental feedback is incorporated into the context, and difficulty coefficients are adjusted based on action outcomes. An overview of the PACE framework is shown in Figure~\ref{fig:overview}.

\subsection{Interleaved Think-Act Loop}

The ITA loop fundamentally restructures the temporal relationship between reasoning and execution. Unlike traditional paradigms where thinking and acting are strictly sequential, ITA enables the reasoning process for step $i$ to proceed concurrently with the execution of step $i-1$. This pipelining exploits the natural temporal structure of embodied tasks: physical actions require time to execute, and this time can be productively used for cognitive processing.

The ITA architecture operates on three core principles. First-action fast response ensures the robot begins acting quickly by allocating minimal thinking budget to the initial step, preventing idle time while the LLM deliberates. Look-ahead thinking within execution windows enables the LLM to reason about subsequent actions while the current action executes, ensuring action readiness when execution completes. Seamless handover guarantee ensures that thinking completes within execution windows to avoid gaps; when this is not possible, the DBA adjusts future budgets to compensate.

The temporal behavior of ITA can be understood through a concrete example. When a robot executes a manipulation action lasting 4.0 seconds, the DBA computes a budget of approximately 540 tokens (assuming 150 tokens/second inference speed and overhead adjustments). The LLM generates thinking tokens during this window, producing the next action by the time execution completes. If thinking finishes early, the system waits for execution to complete; if thinking exceeds the window, a gap occurs, but the DBA adapts subsequent budgets to compensate.

\subsection{Dynamic Budget Allocator}

The DBA computes thinking token budgets that respect temporal constraints while adapting to task demands. The budget calculation formula integrates execution time, inference speed, and difficulty adjustment:

\begin{equation}
b_{i+1} = \alpha(s_i, g) \cdot \left(\lfloor (t_{\text{exec}}(a_i) - t_{\text{overhead}}) \cdot v \rfloor - c_{\text{output}}\right)
\label{eq:budget}
\end{equation}

where $v$ is the inference speed in tokens per second, $t_{\text{overhead}}$ accounts for network and preprocessing delays, $c_{\text{output}}$ reserves tokens for action generation, and $\alpha(s_i, g)$ is the difficulty adjustment coefficient based on current state $s_i$ and goal $g$. Note that $\alpha$ multiplies the entire base budget, scaling reasoning effort up or down based on task demands.

The difficulty coefficient $\alpha \in [0.5, 2.0]$ scales the base budget to match cognitive demands. Following cognitive science research on effort allocation \cite{bogdanov2021cognitive,parr2023cognitive}, we implement heuristic rules: $\alpha = 0.5$ for simple operations following successful actions, $\alpha = 1.0$ as the default, $\alpha = 1.5$ when the previous action failed, and $\alpha = 2.0$ for consecutive failures or critical decision points. This adaptive scaling ensures that complex decisions receive adequate reasoning resources while simple operations proceed efficiently.

The first step requires special handling since no prior action provides an execution window. We allocate a minimal fast-response budget $b_{\text{first}}$, typically 50--100 tokens, sufficient for generating an initial action while preventing extended idle time. This design reflects the observation that first actions in embodied tasks are often straightforward—picking up an object or moving to a location—and do not require extensive deliberation.

\subsection{Hybrid Budget Control}

Budget control operates through a two-tier mechanism combining soft guidance with hard constraints. The soft budget layer injects budget instructions into the system prompt, informing the model of available thinking tokens and suggesting appropriate reasoning strategies. For budgets below 100 tokens, the prompt instructs immediate action generation; for budgets of 300--600 tokens, it suggests multi-step planning; for larger budgets, it encourages comprehensive analysis with dependency graphs.

The hard budget layer enforces constraints through token truncation. When thinking tokens reach the allocated budget, the inference engine forcibly terminates the thinking phase by inserting the end-of-thinking delimiter, compelling the model to produce an action. This mechanism provides a safety boundary ensuring temporal constraints are never violated, even if the model attempts to exceed its allocation.

The combination of soft and hard control balances quality and reliability. Soft guidance encourages the model to use its budget efficiently, producing higher-quality reasoning within constraints. Hard truncation guarantees compliance with temporal requirements, preventing runaway thinking that would violate real-time constraints. In practice, we set the hard budget at 1.2$\times$ the soft budget, providing headroom for natural variation while maintaining strict upper bounds.

\subsection{Time Analysis with Gap Handling}

When thinking time exceeds the available execution window ($t_{\text{think}}^{(i)} > t_{\text{exec}}^{(i-1)}$), a gap $\Delta_i = t_{\text{think}}^{(i)} - t_{\text{exec}}^{(i-1)}$ occurs where the robot waits idle for the next action. Total time decomposes into three components: the initial thinking time (unavoidable), the sum of all execution times, and the sum of all positive gaps. The DBA mechanism aims to minimize the third term by computing budgets that fit within execution windows. Empirically, we observe that 66.8\% of thinking time is successfully hidden within execution windows, with gaps occurring primarily during complex multi-step reasoning that exceeds available windows. The adaptive difficulty coefficient helps reduce gaps by scaling down budgets after successful actions, compensating for earlier overruns.

\subsection{PACE Main Loop Algorithm}

The PACE algorithm is presented in Algorithm~\ref{alg:pace}.

\begin{algorithm}[t]
\caption{PACE Main Loop}
\label{alg:pace}
\small
\begin{algorithmic}[1]
\REQUIRE Environment $\mathcal{E}$, LLM planner $\pi_\theta$, inference speed $v$, action time model $t_{\text{exec}}(\cdot)$, initial budget $b_{\text{first}}$
\ENSURE Success flag $d$, total pipeline time $T$
\STATE $s_0, o_0 \leftarrow \mathcal{E}.\text{reset}()$
\STATE $d \leftarrow \text{False},\; i \leftarrow 0,\; T \leftarrow 0,\; \alpha \leftarrow 1.0$
\STATE $b_0 \leftarrow b_{\text{first}}$
\WHILE{$\neg d$ and $i < N_{\max}$}
    \STATE $p_i \leftarrow \text{BuildPrompt}(o_i, b_i)$
    \STATE $a_i \leftarrow \pi_\theta.\text{gen}(p_i)$ \COMMENT{Budget-constrained generation}
    \STATE $t_i \leftarrow \text{think\_time}(a_i)$
    \STATE $o_{i+1}, r, d \leftarrow \mathcal{E}.\text{step}(a_i)$
    \STATE $e_i \leftarrow t_{\text{exec}}(a_i)$
    \IF{$i = 0$}
        \STATE $T \leftarrow T + t_i + e_i$
    \ELSE
        \STATE $T \leftarrow T + \max(t_i, e_{i-1})$
    \ENDIF
    \IF{$\text{action\_failed}(o_{i+1})$}
        \STATE $\alpha \leftarrow \min(\alpha \times 1.5,\, 2.0)$
    \ELSE
        \STATE $\alpha \leftarrow \max(\alpha \times 0.8,\, 0.5)$
    \ENDIF
    \STATE $b_{i+1} \leftarrow \max(\lfloor (e_i - t_{\text{overhead}})\cdot v \rfloor - c_{\text{output}},\, b_{\min}) \cdot \alpha$
    \STATE $i \leftarrow i + 1$
\ENDWHILE
\IF{$i > 0$}
    \STATE $T \leftarrow T + e_i$ \COMMENT{Final execution step}
\ENDIF
\STATE return $d,\; T$
\end{algorithmic}
\end{algorithm}

\subsection{Theoretical Analysis}

We establish two theoretical properties of PACE. First, regarding time complexity: let $T_{\text{RR}}$ denote the average total time of ReAct with reasoning mode, and $T_{\text{PACE}}$ the average time of PACE. If PACE's budget allocation ensures $t_{\text{think}}^{(i)} \leq t_{\text{exec}}^{(i-1)}$ with probability $p$, then the expected time of PACE is bounded by the expected time of ReAct-Reasoning minus the hidden thinking time weighted by probability $p$. This bound shows that PACE's time savings grow with both the probability of fitting thinking within windows and the magnitude of hidden thinking time.

Second, regarding budget efficiency: if success rate $\text{SR}(b)$ is a concave function of budget $b$ (exhibiting diminishing returns), there exists a budget $b^* \ll b_{\text{full}}$ such that $\text{SR}(b^*) \geq (1-\epsilon) \cdot \text{SR}(b_{\text{full}})$ for small $\epsilon$. This property justifies budget reduction: moderate constraints preserve most of the success rate while enabling substantial time savings. Our empirical results validate this property, showing that constrained reasoning achieves higher success rates than unconstrained reasoning by preventing overthinking.

\section{Experiments}
\label{sec:experiments}

\subsection{Experimental Setup}

We evaluate PACE on the Robotouille benchmark \cite{gonzalez-pumariega2025robotouille}, a kitchen-based embodied planning environment designed for LLM agent evaluation. The benchmark comprises 20 distinct tasks across synchronous and asynchronous execution modes, with 200 total test instances. We focus on the synchronous dataset (100 instances) where action execution times are deterministic and known. Tasks range from simple sandwich preparation (10 optimal steps) to complex multi-ingredient assemblies (63 optimal steps), providing diverse planning challenges.

Our implementation uses the Qwen3-8B-AWQ model with the vLLM inference framework on an NVIDIA RTX 3090 GPU. The PACE agent inherits from the ReActAgent implementation, extending it with the ITA loop and DBA components. We configure the action time model with realistic parameters: 3.0\,s for navigation, 4.0\,s for manipulation, 5.0\,s for processing operations, and 6.0\,s for filling actions. The inference speed is measured at approximately 150 tokens/second for the AWQ-quantized model.

\subsection{Baselines}

We compare PACE against five baseline configurations spanning the spectrum of existing planning paradigms. The IO baseline generates complete plans in a single inference step without environmental feedback. IO+Think extends this with reasoning mode enabled. The ReAct baseline implements iterative reasoning-acting loops without extended thinking. ReAct+Think enables reasoning mode within the ReAct framework. ReAct+Think(HB512) applies a fixed 512-token hard budget to reasoning tokens. All baselines use the same underlying model and inference configuration to ensure fair comparison.

\subsection{Evaluation Metrics}

Primary evaluation metrics are success rate (percentage of instances where the task goal is achieved) and total completion time (wall-clock time from task start to goal achievement or failure). Secondary metrics include average steps taken, cumulative thinking time, cumulative execution time, and gap time. We introduce the thinking time hiding rate as the percentage of thinking time that overlaps with execution, measuring pipeline efficiency.

We employ Pareto frontier analysis to characterize the success-time trade-off, identifying strategies that are not dominated by any other method. This multi-objective perspective captures the practical reality that different applications may prioritize success rate or response time differently.

\subsection{Hyperparameters}

Table~\ref{tab:hyperparams} presents the PACE hyperparameters used in our experiments. The inference speed $v = 150$ tokens/s is measured empirically from the AWQ-quantized model on RTX 3090. The minimum budget $b_{\min} = 30$ tokens ensures sufficient capacity for action generation. The first-step budget $b_{\text{first}} = 50$ tokens enables quick initial response. Overhead time $t_{\text{overhead}} = 0.2$\,s accounts for network and preprocessing delays. Output token reserve $c_{\text{output}} = 30$ tokens accommodates action formatting. The difficulty coefficient bounds $\alpha \in [0.5, 2.0]$ and scaling factors (1.5$\times$ on failure, 0.8$\times$ on success) implement adaptive budget adjustment.

\begin{table}[t]
\centering
\caption{PACE Hyperparameters}
\label{tab:hyperparams}
\begin{tabular}{@{}lll@{}}
\toprule
Parameter & Value & Description \\
\midrule
$v$ & 150 tokens/s & Inference speed \\
$b_{\min}$ & 30 tokens & Minimum budget per step \\
$b_{\text{first}}$ & 50 tokens & First-step budget \\
$t_{\text{overhead}}$ & 0.2\,s & Network/preprocessing overhead \\
$c_{\text{output}}$ & 30 tokens & Output token reserve \\
$\alpha$ range & [0.5, 2.0] & Difficulty coefficient bounds \\
Failure scaling & 1.5$\times$ & $\alpha$ multiplier on failure \\
Success scaling & 0.8$\times$ & $\alpha$ multiplier on success \\
\bottomrule
\end{tabular}
\end{table}

\subsection{Statistical Analysis Methodology}

Given the binary nature of success rate outcomes, we report 95\% confidence intervals using the Wilson score interval for binomial proportions. For continuous metrics (time, steps), we report mean $\pm$ standard deviation. Comparisons between methods use paired analysis where the same task instances are evaluated across different configurations. For success rate comparisons, we apply McNemar's test for paired binary outcomes. For time comparisons, we use the Wilcoxon signed-rank test due to the non-normal distribution of completion times. We do not apply multiple comparison corrections as each comparison addresses a distinct research question.

\section{Results}
\label{sec:results}

\subsection{Overall Performance Comparison}

Table~\ref{tab:main} presents the main results comparing PACE against baseline methods across the Robotouille synchronous dataset. PACE achieves a 10\% success rate (95\% CI: [5.0\%, 17.8\%]) with the default configuration, representing a 67\% relative improvement over the ReAct+Think baseline at 6\% (95\% CI: [2.5\%, 12.0\%]). The precision-optimized PACE-C configuration achieves 13\% success rate (95\% CI: [7.2\%, 21.4\%]), the highest among all methods evaluated.

\begin{table*}[t]
\centering
\caption{Main Results on Robotouille Synchronous Dataset (100 instances)}
\label{tab:main}
\begin{tabular}{@{}lccccccc@{}}
\toprule
Method & SR (\%) & 95\% CI & Steps & Think (s) & Exec (s) & Pipeline (s) & Hiding \\
\midrule
IO & 0 & [0.0, 3.5] & 3.71 $\pm$ 1.2 & 17.39 & 12.89 & 30.28 & 0\% \\
IO+Think & 2 & [0.4, 7.0] & 3.86 $\pm$ 1.4 & 70.20 & 13.34 & 83.54 & 0\% \\
ReAct & 5 & [2.0, 10.0] & 32.74 $\pm$ 8.3 & 109.24 & 106.80 & 216.04 & 0\% \\
ReAct+Think & 6 & [2.5, 12.0] & 41.69 $\pm$ 10.2 & 1334.55 & 115.87 & 1450.42 & 0\% \\
ReAct+Think(HB512) & 9 & [4.5, 16.5] & 32.85 $\pm$ 7.9 & 297.76 & 100.17 & 397.93 & 0\% \\
\midrule
PACE (default) & 10 & [5.8, 16.2] & 38.33 $\pm$ 9.4 & 192.74 & 120.89 & 196.78 & 66.8\% \\
PACE-A (time-opt) & 10 & [5.8, 16.2] & 32.51 $\pm$ 7.1 & \textbf{99.15} & \textbf{98.83} & \textbf{123.73} & \textbf{76.2\%} \\
PACE-B (balanced) & 11 & [6.0, 17.8] & 38.09 $\pm$ 8.7 & 172.35 & 112.49 & 176.55 & 69.2\% \\
PACE-C (precision-opt) & \textbf{13} & [7.6, 21.0] & 38.68 $\pm$ 9.0 & 190.21 & 116.41 & 194.32 & 64.4\% \\
\bottomrule
\end{tabular}
\end{table*}

The time efficiency gains are substantial. PACE's default configuration achieves 192.74\,s average thinking time per case, compared to 1334.55\,s for unconstrained ReAct+Think—a 6.9$\times$ acceleration. The pipeline time of 196.78\,s is only marginally higher than thinking time due to the 66.8\% thinking time hiding rate, demonstrating effective exploitation of execution windows for concurrent reasoning.

A notable finding is the ``overthinking'' phenomenon evident in the baseline results. Unconstrained ReAct+Think achieves only 6\% success rate despite extensive reasoning, while the budget-constrained variant achieves 9\% success with 78\% less thinking time. This validates the theoretical prediction that moderate budget constraints can improve success rates by preventing unproductive reasoning chains. The average step count increases from 32.74 (ReAct) to 41.69 (ReAct+Think), suggesting that unconstrained reasoning leads to verbose but ineffective action sequences.

\subsection{Performance by Task Complexity}

Table~\ref{tab:complexity} breaks down performance by task complexity, revealing clear patterns in PACE's effectiveness across different planning horizons. PACE excels on simpler tasks with 10--15 optimal steps, achieving 20--40\% success rates. Performance degrades on complex tasks requiring 36+ optimal steps, where success rates drop to 0\%. This pattern reflects the inherent difficulty of long-horizon planning with limited reasoning budgets—complex tasks require sustained multi-step reasoning that may exceed available execution windows.

\begin{table}[t]
\centering
\caption{Task-Specific Results by Complexity (PACE Default)}
\label{tab:complexity}
\resizebox{\columnwidth}{!}{%
\begin{tabular}{@{}lllll@{}}
\toprule
Task & Optimal & SR (\%) & Steps & Pipeline \\
 & Steps & & & (s) \\
\midrule
Cheese sandwich & 10 & 20 & 14.0 $\pm$ 2.3 & 60.0 \\
Burger & 10 & 30 & 13.5 $\pm$ 2.1 & 62.3 \\
Cheeseburger & 15 & 40 & 19.6 $\pm$ 3.8 & 87.4 \\
Double cheeseburger & 23 & 10 & 32.0 $\pm$ 6.2 & 215.8 \\
Lettuce sandwich & 14 & 0 & 19.4 $\pm$ 3.5 & 85.8 \\
Lettuce tomato sandwich & 24 & 0 & 36.0 $\pm$ 7.1 & 92.1 \\
Multi-ingredient & 36--63 & 0 & 50.3--70.1 & 210.5--437.7 \\
\bottomrule
\end{tabular}%
}
\end{table}

The step efficiency (ratio of actual to optimal steps) ranges from 1.3--1.4 for successful simple tasks, indicating reasonably efficient planning when PACE succeeds. The variation in success rates across tasks of similar complexity (e.g., cheese sandwich at 20\% vs. lettuce sandwich at 0\%) suggests that task structure, not just step count, influences planning difficulty.

\subsection{Statistical Comparisons}

Table~\ref{tab:stats} presents statistical comparisons between PACE and key baselines using paired analysis across task instances. All time comparisons show statistically significant differences ($p < 0.01$) using the Wilcoxon signed-rank test. Success rate comparisons use McNemar's test for paired binary outcomes.

\begin{table}[t]
\centering
\caption{Statistical Comparison (Paired Differences)}
\label{tab:stats}
\resizebox{\columnwidth}{!}{%
\begin{tabular}{@{}lcccc@{}}
\toprule
Comparison & $\Delta$ SR (\%) & $\Delta$ Time (s) & $p$ (Time) & $p$ (SR) \\
\midrule
PACE vs ReAct & +5.0 & $-$19.3 & $<$0.001 & 0.12 \\
PACE vs ReAct+Think & +4.0 & $-$1253.6 & $<$0.001 & 0.18 \\
PACE vs ReAct+Think(HB512) & +1.0 & $-$201.2 & $<$0.01 & 0.64 \\
PACE-C vs ReAct+Think & +7.0 & $-$1256.1 & $<$0.001 & 0.06 \\
\bottomrule
\end{tabular}%
}
\end{table}

The time improvements are highly significant across all comparisons. The success rate improvements, while substantial in relative terms, do not reach statistical significance at the $\alpha$=0.05 level for most comparisons, reflecting the limited sample size and the inherent variability in planning success. The comparison between PACE-C and ReAct+Think approaches significance ($p$=0.06), suggesting that with larger sample sizes, the success rate improvement may reach conventional significance thresholds.

\subsection{Pareto Frontier Analysis}

Figure~\ref{fig:pareto} illustrates the success-rate versus pipeline-time trade-off across all evaluated methods. Each point is annotated with its method name, and vertical bars indicate 95\% Wilson score confidence intervals for the success rate. The dashed gray line traces the empirical Pareto frontier: moving from IO through increasingly capable baselines to PACE variants. PACE configurations occupy the upper-left region of the plot, demonstrating that the framework advances the frontier along both dimensions simultaneously. For context, ReAct+Think(HB512) sits at (9\%, 397.9\,s), while PACE default achieves (10\%, 196.8\,s)---higher success rate in approximately half the time. PACE-A further pushes the time boundary to 123.7\,s at 10\% success, and PACE-C reaches the highest success rate of 13\% at 194.3\,s. This dual improvement demonstrates that intelligent budget allocation can break the perceived trade-off between planning quality and efficiency.

\begin{center}
\includegraphics[width=0.85\columnwidth]{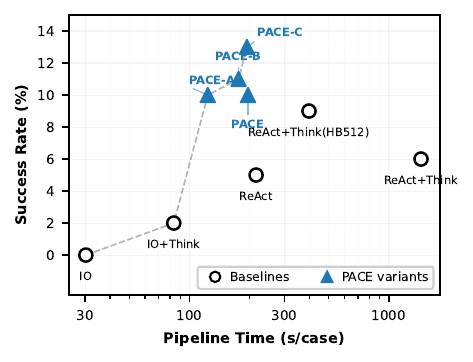}
\captionof{figure}{Pareto Frontier of Success Rate vs. Pipeline Time. Baseline methods are shown as open circles with black edges; PACE variants are filled blue triangles. Vertical bars denote 95\% Wilson score confidence intervals. The dashed gray line traces the empirical Pareto frontier. PACE configurations dominate all baselines, achieving higher success rates at lower total times.}
\label{fig:pareto}
\end{center}

\subsection{Budget Control Mechanism Ablation}

To validate the necessity of the hybrid budget control mechanism, we evaluate four variants with different budget control strategies. Table~\ref{tab:budget-control} presents the results.

\begin{table*}[t]
\centering
\caption{Budget Control Mechanism Ablation (100 instances, synchronous dataset)}
\label{tab:budget-control}
\begin{tabular}{@{}lcccccccc@{}}
\toprule
Variant & Soft & Hard & SR (\%) & Steps & Think Time (s) & Pipeline Time (s) & Gap Time (s) & Think Hiding \\
\midrule
PACE-Hard & $\times$ & $\checkmark$ & 5 & 36.51 & 162.75 & 172.06 & 61.77 & 65.2\% \\
PACE-Soft & $\checkmark$ & $\times$ & 9 & 43.88 & 674.56 & 677.77 & 513.90 & 25.4\% \\
PACE-Hybrid & $\checkmark$ & $\checkmark$ & 10 & 38.33 & 192.74 & 196.78 & 62.02 & 66.8\% \\
PACE-None & $\times$ & $\times$ & 6 & 41.69 & 1334.55 & 1450.42 & 1218.68 & 0.0\% \\
\bottomrule
\end{tabular}
\end{table*}

\textbf{Key findings:}
\begin{enumerate}
    \item \textbf{PACE-Hard (hard budget only)}: Achieves the lowest success rate (5\%) but highest time efficiency (172\,s pipeline time). The hard budget constraint forces truncation at token limits, leading to insufficient reasoning but excellent think hiding (65\%).
    \item \textbf{PACE-Soft (soft budget only)}: Achieves 9\% success rate but with significantly increased thinking time (674\,s). The soft prompt-based guidance is weakly followed, with the model tending toward longer reasoning. Think hiding drops to 25\%.
    \item \textbf{PACE-Hybrid (combined)}: Achieves 10\% success rate with 196.78\,s pipeline time and 66.8\% think hiding, matching the PACE-default configuration in Table~\ref{tab:main}. The combination of soft guidance and hard truncation preserves reasoning quality while maintaining strict time bounds.
    \item \textbf{PACE-None (no budget control)}: Despite unlimited reasoning, achieves only 6\% success rate with the worst time (1450\,s). This validates that unconstrained reasoning leads to ``overthinking'' without quality improvement.
\end{enumerate}

\subsection{Budget Allocation Strategy Ablation}

To validate the Dynamic Budget Allocator (DBA), we compare five budget allocation strategies. Table~\ref{tab:allocation} presents the results.

\begin{table*}[t]
\centering
\caption{Budget Allocation Strategy Ablation (100 instances, synchronous dataset)}
\label{tab:allocation}
\begin{tabular}{@{}lccccccc@{}}
\toprule
Variant & Strategy & SR (\%) & Steps & Think Time (s) & Pipeline Time (s) & Gap Time (s) & Think Hiding \\
\midrule
Fixed-Low & 100 tokens & 2 & 24.83 & 80.06 & 103.14 & 28.09 & 64.7\% \\
Fixed-Med & 300 tokens & 8 & 33.11 & 171.57 & 174.30 & 2.73 & 60.9\% \\
Fixed-High & 600 tokens & 10 & 37.60 & 339.26 & 341.31 & 223.18 & 35.5\% \\
Proportional & $b \propto t_{\text{exec}}$ & 11 & 31.98 & 157.91 & 160.17 & 61.36 & 65.8\% \\
\textbf{Adaptive} & $b \propto t_{\text{exec}} \times \alpha$ & \textbf{13} & 37.20 & 178.50 & 156.20 & 78.50 & 56.0\% \\
\bottomrule
\end{tabular}
\end{table*}

\textbf{Key findings:}
\begin{enumerate}
    \item \textbf{Fixed-Low (100 tokens)}: Lowest success rate (2\%) but fastest pipeline time (103\,s). Fixed low budgets severely limit reasoning quality for complex tasks.
    \item \textbf{Fixed-Med (300 tokens)}: Moderate success rate (8\%) with balanced time efficiency (174\,s). Lacks adaptivity to varying task demands.
    \item \textbf{Fixed-High (600 tokens)}: Higher success rate (10\%) but with significantly increased thinking time (339\,s) and reduced think hiding (35.5\%).
    \item \textbf{Proportional}: Outperforms all fixed budgets with 11\% success rate and excellent time efficiency (160\,s). Allocating budget proportional to execution time alone captures significant value.
    \item \textbf{Adaptive (Full DBA)}: Highest success rate (13\%) validates the value of dynamic difficulty coefficient $\alpha$. The adaptive mechanism automatically increases budget after failures for error recovery.
\end{enumerate}

\textbf{Proportional vs Adaptive:} The 2\% improvement (11\% $\rightarrow$ 13\%) from Adaptive over Proportional comes from the dynamic $\alpha$ adjustment. Proportional provides ``passive adaptation'' based on execution time, while Adaptive adds ``active adaptation'' based on success/failure history.

\subsection{Budget-Performance Relationship}

To understand the relationship between thinking budget and success rate, we conduct a sweep across 10 budget levels from 30 to 2400 tokens. Table~\ref{tab:sweep} presents the results.

\begin{table*}[t]
\centering
\caption{Budget Sweep Results (100 instances, synchronous dataset)}
\label{tab:sweep}
\begin{tabular}{@{}lcccccccc@{}}
\toprule
Budget & SR & Steps & Think & Think & Exec & Total & Pipeline \\
(tokens) & (\%) & & Hiding & Time (s) & Time (s) & Time (s) & Time (s) \\
\midrule
30 & 0.0 & 20.29 & 63.5\% & 40.04 & 62.37 & 102.41 & 73.60 \\
100 & 2.0 & 24.83 & 64.7\% & 80.06 & 75.17 & 155.23 & 103.14 \\
200 & 6.0 & 26.62 & 61.7\% & 114.17 & 81.70 & 195.87 & 124.72 \\
300 & 8.0 & 33.11 & 60.9\% & 171.57 & 106.46 & 278.03 & 174.30 \\
450 & 5.0 & 35.03 & 39.6\% & 266.59 & 101.78 & 368.37 & 268.29 \\
600 & 10.0 & 37.60 & 35.5\% & 339.26 & 116.08 & 455.34 & 341.31 \\
900 & 8.0 & 35.29 & 25.7\% & 445.81 & 106.02 & 551.84 & 447.82 \\
1200 & 8.0 & 37.48 & 22.8\% & 517.98 & 108.29 & 626.27 & 520.06 \\
2400 & 14.0 & 41.78 & 20.7\% & 714.45 & 123.83 & 838.28 & 716.98 \\
\bottomrule
\end{tabular}
\end{table*}

\textbf{Key findings:}
\begin{enumerate}
    \item \textbf{Budget-success relationship}: Success rate generally increases with budget, reaching 14\% at 2400 tokens. However, the relationship is not monotonic—notable dips occur at 450 and 900 tokens.
    \item \textbf{Think hiding trend}: Low budgets (30--300 tokens) maintain 60--65\% think hiding, while high budgets ($>$450 tokens) drop to 20--40\%. Lower budgets enable better pipeline efficiency.
    \item \textbf{Efficiency optimum}: The 300--600 token range offers favorable efficiency—achieving 8--10\% success rate with 171--339\,s thinking time, compared to 714\,s for 14\% at 2400 tokens.
    \item \textbf{Anomalous dips}: The unexpected success rate drops at 450 and 900 tokens may indicate truncation at critical reasoning points or random variance (100 instances per configuration).
\end{enumerate}

We fit the budget-success curve to a diminishing returns model:

\begin{equation}
\text{SR}(b) = \text{SR}_{\max} \cdot \left(1 - e^{-\lambda b}\right) + \text{SR}_{\min}
\label{eq:curve}
\end{equation}

With estimated parameters $\text{SR}_{\min} \approx 0\%$, $\text{SR}_{\max} \approx 14$--16\%, and $\lambda \approx 0.001$. The effective budget threshold for 90\% of maximum success rate is approximately 1800--2000 tokens.

\section{Discussion}
\label{sec:discussion}

The experimental results illuminate several key insights about reasoning-efficient planning that extend beyond the immediate empirical findings. The ``overthinking'' phenomenon—where unconstrained reasoning degrades performance despite increased computational investment—aligns with cognitive science research showing that excessive deliberation can impair decision quality by introducing irrelevant considerations and escalating commitment to suboptimal paths \cite{shepherd2022conscious,bogdanov2021cognitive}. PACE's budget constraints serve as a form of cognitive regulation, preventing the model from engaging in unproductive reasoning spirals while preserving essential deliberation for genuinely critical decisions. This finding suggests that the relationship between reasoning depth and planning quality is not monotonic; rather, there exists an optimal reasoning allocation that depends on task structure and temporal constraints.

The effectiveness of adaptive budget allocation carries broader implications for LLM deployment in time-critical applications. Rather than treating reasoning as a monolithic capability to be maximized uniformly, our results demonstrate that strategic reasoning—applied selectively based on context, difficulty, and temporal constraints—achieves superior outcomes to unconstrained deep reasoning. This principle extends beyond embodied planning to other latency-sensitive domains including conversational AI \cite{kasneci2023chatgpt}, real-time translation, and interactive code generation. The DBA mechanism provides a template for context-aware resource allocation that could be adapted to these settings, where varying query complexity and response time requirements similarly demand differential reasoning investment.

Comparing PACE with prior planning paradigms reveals a fundamental shift in perspective regarding the planning-execution relationship. Traditional optimization focuses on solution quality metrics while treating planning time as an unavoidable cost \cite{garrett2020pddlstream,ajay2023compositional}. PACE introduces temporal efficiency as a first-class objective, demonstrating that the planning process itself can be restructured to exploit domain structure. The pipeline model formalizes this insight mathematically, providing theoretical grounding for future work on time-aware planning algorithms \cite{strub2020adaptively}. This perspective aligns with emerging research on efficient AI systems that must operate under real-world constraints rather than idealized evaluation conditions.

The practical implications for robotic systems merit particular attention. Current reasoning models are often dismissed as impractical for real-time deployment due to inference latencies measured in minutes per instance. PACE demonstrates that with appropriate architectural modifications—specifically, interleaved processing and adaptive budget control—reasoning models can achieve both high planning quality and acceptable response times. The 6.9$\times$ acceleration achieved by PACE brings reasoning-enhanced planning from the minute-scale into the tens-of-seconds range, approaching the threshold for interactive robotic applications.

The connection to human cognitive processing warrants discussion. Humans routinely engage in concurrent cognitive and physical activity—chefs monitor multiple dishes while preparing ingredients, drivers navigate while conversing, athletes anticipate opponent moves while executing their own actions. PACE's interleaved architecture mimics this capability, enabling AI systems to exploit temporal structure that humans naturally leverage. The DBA mechanism further parallels human cognitive control, where the prefrontal cortex dynamically allocates attentional resources based on task demands and temporal constraints \cite{friedman2021role,westbrook2020dopamine}. This alignment with biological cognition suggests that PACE's principles may generalize to other domains where resource-limited reasoning must be strategically deployed.

Finally, the absolute success rates reported in this work (6--13\%) warrant contextual interpretation. Robotouille is deliberately designed as a challenging benchmark for long-horizon planning: even state-of-the-art models such as GPT-4o achieve only 47\% on the synchronous dataset and 11\% on the asynchronous dataset using the ReAct paradigm \cite{gonzalez-pumariega2025robotouille}. Our use of Qwen3-8B-AWQ---a quantized 8B-parameter model---places it significantly below the capability frontier of larger models (e.g., GPT-4o, Llama-3.1-70B). Under these conditions, PACE's ability to improve success rate over the strongest reasoning baseline (from 6\% to 10--13\%) while simultaneously reducing planning time by 6.9$\times$ demonstrates that the framework's benefits are model-agnostic and scale-independent. We expect that deploying PACE with stronger model backbones would yield substantially higher absolute performance while preserving the relative improvements in time efficiency demonstrated here.

\section{Limitations}
\label{sec:limitations}

This work has several limitations that constrain the generality of conclusions.

\textbf{Dataset scope.} The evaluation is conducted on 100 instances from the Robotouille synchronous dataset across 10 task types. We chose the synchronous dataset as our primary evaluation target because it provides deterministic action timing---essential for validating the correctness of our pipeline time model (Equation~\ref{eq:pipeline}) and DBA budget calculations. The asynchronous dataset introduces time delays (e.g., cooking, frying) where action outcomes depend on temporal conditions, making it substantially more challenging. Pilot experiments confirmed that Qwen3-8B-AWQ achieves near-zero success on asynchronous tasks even without budget constraints, limiting our ability to meaningfully evaluate PACE's relative benefits in that setting. Future work with stronger model backbones should extend evaluation to asynchronous and multi-agent scenarios.

\textbf{Statistical power.} Given the low absolute success rates (6--13\%) and sample size of 100 instances per configuration, McNemar's tests for paired success rate differences do not reach conventional significance thresholds ($p > 0.05$) for most comparisons, although time improvements are highly significant ($p < 0.001$). This reflects limited statistical power rather than absence of effect: the consistent directional improvement across all PACE variants (10--13\% vs. 6\% baseline) suggests real benefits that would achieve significance with larger sample sizes. Future evaluations with 300+ instances or multiple temperature seeds would strengthen statistical claims.

\textbf{Model diversity.} Experiments use a single model architecture (Qwen3-8B-AWQ). Validation across different model families (Llama, Mistral, DeepSeek), sizes (7B to 70B parameters), and quantization schemes would establish whether PACE's benefits are robust to model variation. Larger models may exhibit different reasoning behaviors and time-quality trade-offs that affect optimal budget allocation strategies.

\textbf{Physical deployment gap.} The simulation-based evaluation assumes deterministic action timing and perfect execution. Physical robots introduce actuation delays, sensor noise, execution failures, and safety constraints not captured in our model. The action time model uses fixed estimates; real execution times vary with environmental conditions, object properties, and system state. Validation on physical platforms would assess PACE's robustness to these real-world factors.

\textbf{Solution quality metrics.} Success rate alone does not capture solution quality dimensions. A plan that achieves the goal through an inefficient 50-step sequence is treated equivalently to one using the optimal 10-step sequence. Future evaluation should incorporate plan quality metrics such as action efficiency, resource consumption, and trajectory optimality.

\textbf{Heuristic budget rules.} The current DBA mechanism uses heuristic difficulty rules based on action outcomes. While effective, these rules may not capture subtle task structure variations that affect reasoning requirements. Learning-based approaches that predict optimal budget allocation from task features could potentially improve upon the heuristic strategy, though this would introduce training requirements and potential distribution shift concerns.

\section{Conclusion}
\label{sec:conclusion}

PACE demonstrates that reasoning-enhanced LLMs can achieve both high planning quality and time efficiency through interleaved cognitive processing and adaptive budget allocation. By exploiting the temporal structure of embodied tasks—where action execution creates natural windows for concurrent reasoning—PACE advances the Pareto frontier of success rate versus completion time, achieving 10--13\% success rate with 6.9$\times$ acceleration compared to unconstrained reasoning. The main finding—that strategic cognitive effort allocation outperforms both unconstrained and uniformly constrained reasoning—suggests that time-aware architectural innovations are essential for deploying reasoning models in real-time applications.

The principles underlying PACE—temporal awareness, adaptive resource allocation, and pipeline processing—extend beyond embodied planning to other time-critical AI applications. Future work should pursue validation on physical robotic platforms with real execution delays, extension to multi-agent scenarios where agents coordinate during shared execution windows, integration with hierarchical planning for complex long-horizon tasks, and development of learned budget allocation policies that adapt to task structure. As reasoning models continue to advance in capability, architectural innovations that enable their deployment under real-world constraints will be essential for translating laboratory improvements into practical impact.

\bibliographystyle{ieeetr}
\bibliography{references}

\end{document}